\documentclass[letterpaper, 10 pt, conference, twoside]{ieeeconf}  
\pdfoutput=1

\IEEEoverridecommandlockouts                              

\usepackage{amsmath,amssymb,amsfonts}
\usepackage{calrsfs}
\DeclareMathAlphabet{\pazocal}{OMS}{zplm}{m}{n}
\usepackage{cite}
\usepackage{graphicx}
\usepackage{mathrsfs}
\usepackage{xcolor}
\usepackage{balance}

\title{\LARGE \bf
Autonomous Quilt Spreading for Caregiving Robots
}

\author{Yuchun Guo, Zhiqing Lu, Yanling Zhou and Xin Jiang
	\thanks{This work was supported in part by the Stable Support Research Project under Grant GXWD20220811151420001.}
	\thanks{Yuchun Guo, Zhiqing Lu, Yanling Zhou and Xin Jiang are with the Department of Mechanical Engineering, Harbin Institute of Technology, Taoyuan Sub-district, Nanshan District, Shenzhen, 518055, China, {\tt\small 22S053013@stu.hit.edu.cn}}%
	\thanks{*Corresponding author: Xin Jiang, {\tt\small x.jiang@ieee.org}}
}

\begin{document}

\maketitle
\thispagestyle{empty}
\pagestyle{empty}

\begin{abstract}

In this work, we propose a novel strategy to ensure infants, who inadvertently displace their quilts during sleep, are promptly and accurately re-covered. Our approach is formulated into two subsequent steps: interference resolution and quilt spreading. By leveraging the DWPose human skeletal detection and the Segment Anything instance segmentation models, the proposed method can accurately recognize the states of the infant and the quilt over her, which involves addressing the interferences resulted from an infant's limbs laid on part of the quilt. Building upon prior research, the EM*D deep learning model is employed to forecast quilt state transitions before and after quilt spreading actions. To improve the sensitivity of the network in distinguishing state variation of the handled quilt, we introduce an enhanced loss function that translates the voxelized quilt state into a more representative one. Both simulation and real-world experiments validate the efficacy of our method, in spreading and recover a quilt over an infant.

\end{abstract}
\begin{keywords}
Deformable Object Manipulation, Deep Learning in Robotics, Caregiving Robot
\end{keywords}
\section{INTRODUCTION}

\begin{figure}[t!]
	\centering
	\includegraphics[width=\linewidth]{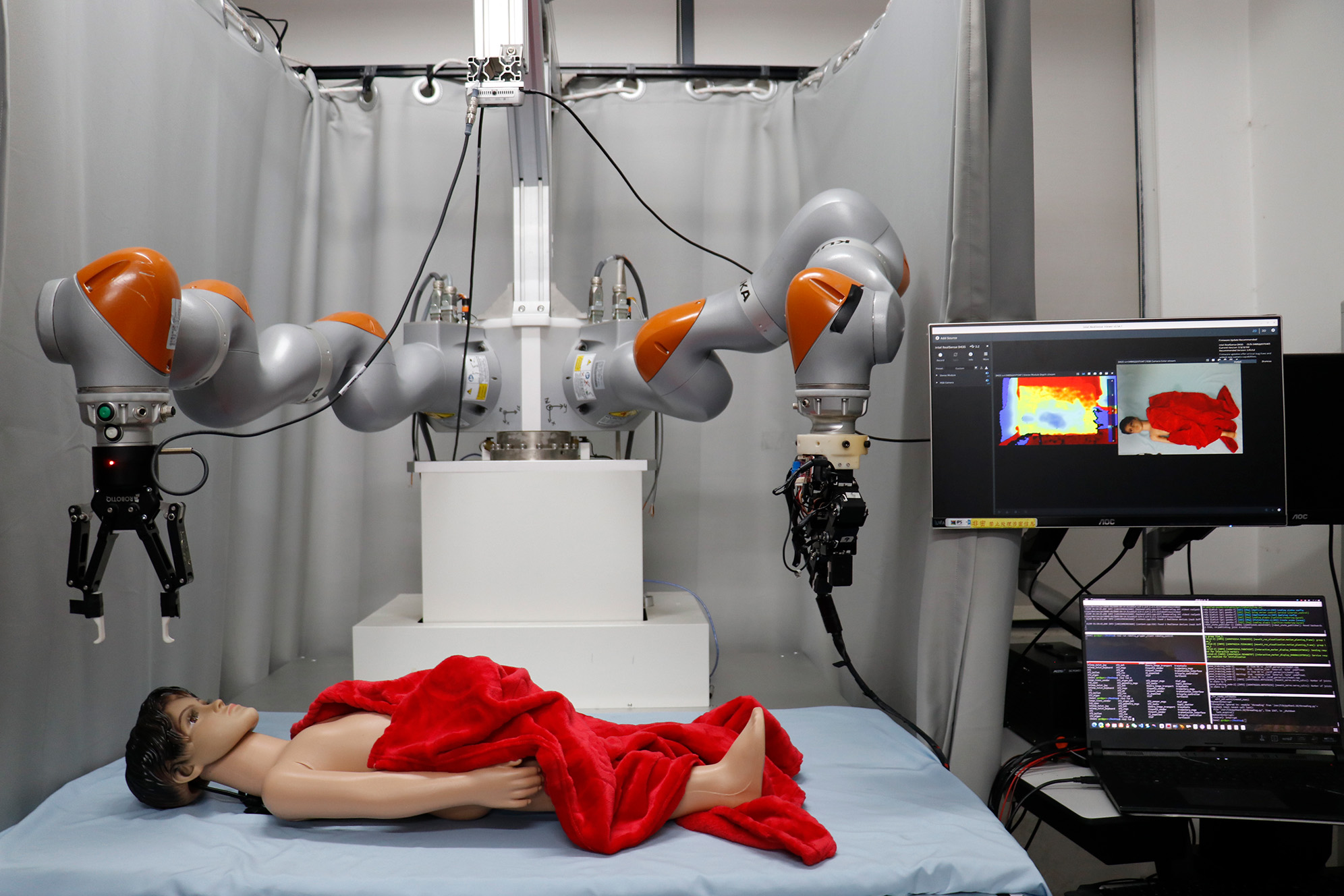}
	\vspace{-2em}
	\caption{Robotic quilt spreading for cover a baby.}
	\label{real}
	\vspace{-2em}
\end{figure}

This work investigates the application of skeletal detection and segmentation techniques, combined with a deep learning model, to efficiently spread a quilt over an infant, addressing challenges posed by limb interference. While robots excel at manipulating rigid objects, handling flexible materials—crucial in textiles \cite{AgeGender, PickingSoft} and medicine \cite{InteractiveSimulation}—remains a challenge. The primary objective of this work is to devise an manipulation actions to ensure infants, especially when their limbs are laid on a quilt during sleep, remain adequately covered (see Fig. \ref{real}).

\subsection{Fabric state recognition}
Accurate state recognition of a deformable object, relying solely on static RGB images, isn't always reliable due to fabric folds. Minor disturbances can change a fabric's state, emphasizing the intricacy of its state space. Recent research leans on active vision based strategies to improve accuracy by reducing fabric state variation space \cite{GarmentPerception, PerceptionCloth, BimanualRobotic}. Researchers employ strategies, such as lifting a point on the fabric in order to reduce randomness of a laid fabric. Visual recognition of fabric typically involves identifying the contours, folds of the fabric, as well as garment features such as collars and cuffs \cite{ClothesFolding, UsingDepth, GarmentRecognition}. In robot assisted dressing, except the state of fabric, the state of human is also important for a smooth interaction. \cite{LearningPut}. High resolution RGB-D camera is widely adopted in capturing the state of a fabric. With it, one can recognize wrinkles then flatten garments \cite{AccurateGarment}. Neural network exhibit promising performance in recognizing fabric states. A well trained deep network model can help to discern crucial grasping regions on fabric such as edges and wrinkles \cite{ClothRegion}. By collecting extensive deformation data of various fabric types within simulators, neural networks can discern and perform tasks across different fabric colors, shapes, textures, and sizes \cite{LearningDense}. Compared to RGB images, tactile sensors can directly capture fabric morphology when they are fixed to the fingertips. Training a classifier in conjunction with these sensors can determine if a robot has grasped a specific number of fabric layers \cite{LearningSingulate}.

\subsection{Fabric Manipulation Planning}
Given the characteristics of robotic end-effectors \cite{GraspingcenteredAnalysis}, there might be differences between the planned effects and actual ones. The selection of grippers should aim to minimize the differences. For tasks such as flattening fabrics, there might not be a need for a specially designed gripper; the fabric can be flattened by sliding it against a flat surface like a table. Most conventional fabric manipulations utilize simple parallel grippers. The planning for fabric manipulation predominantly employs neural networks. An innovative approach is the EM*D (Encode-Manipulate-Decode) Net \cite{FastFlexible}. This network is formulated as a one step forward dynamics of a fabric with manipulation action also embedded within it. When it needs to derive a manipulation action given a desired fabric state, the network will be used in an inverse dynamics style. The corresponding manipulation action is determined by searching with the trained network until the action as the input can result in the desired fabric state. Another research has extended the Visual Foresight framework to introduce VisuoSpatial Foresight (VSF) \cite{VisuospatialMulti}. This method learns fabric dynamics solely in simulation using domain randomized RGB images and depth maps, aiming to achieve versatile fabric manipulation tasks with a single goal-conditioned policy. Additionally, several studies have employed reinforcement learning strategies to learn manipulating soft materials like fabrics in simulation environments \cite{BodiesUncovered, DeepTransfer, LearningManipulate, SimtoReal, VisuospatialForesight, DeepReinforcement, RLClothing, wang2023policy}. Training the aforementioned neural network requires extensive data collection, which can be labor-intensive and costly in practice. Researchers have developed simulators for service robots and cloth manipulation \cite{SoftGym, AssistiveGym}, demonstrating that their simulated outcomes closely resemble real-world scenarios. Except data driven methods, manipulation planning is also possible using a novel vision-based controller for deforming unknown elastic objects, leveraging online estimation and a dynamic-state feedback mechanism \cite{UncalibratedVision}.

\begin{figure}[t!]
	\centering
	\includegraphics[width=\linewidth]{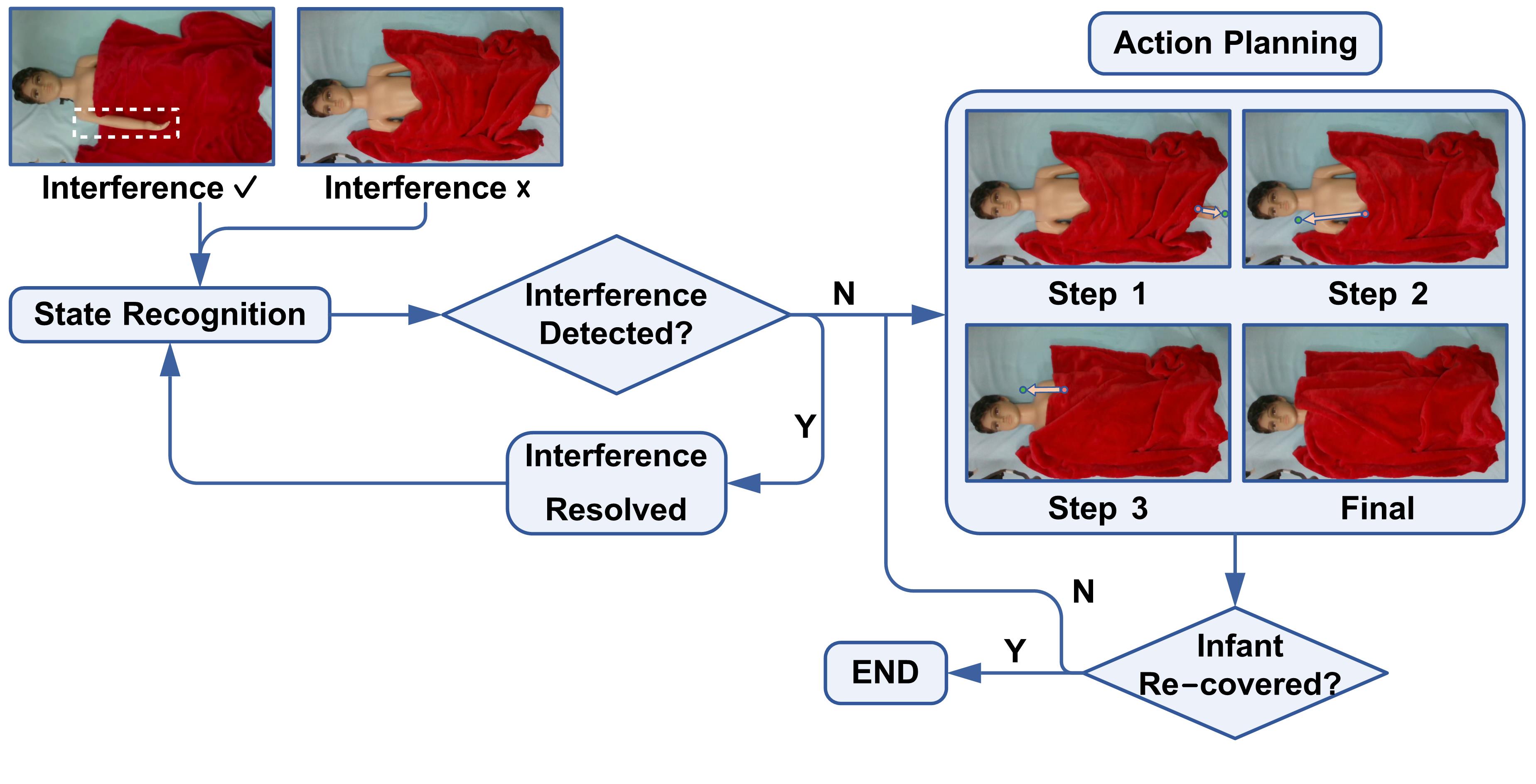}
	\vspace{-2em}
	\caption{The strategy for realizing autonomous quilt spreading.}
	\label{process}
	\vspace{-2em}
\end{figure}

\subsection{Goals, Objectives, and Contributions}
The aim of this paper is to develop an manipulation plan ensuring infants, who inadvertently kick off their quilts during sleep, remain covered, as shown in Fig. \ref{process}. In this strategy, autonomous quilt spreading considered in this paper can be realized by sequentially solving two tasks: interference solving, quilt spreading. The first task is to make sure that the body layer and the quit layer are topologically isolated. After achieving the first task, the subsequent task of quilt spreading for good body coverage is similar to manipulating a free fabric on a table. The key objectives and primary contributions of this work are as follows:
1) A vision-based method for recognizing the state of a quilt considering its topological interference with a human body being covered. 
2) A manipulation strategy for solving the aforementioned interference. 
3) A deep learning-based quilt spreading action planning for realizing a good re-coverage of human body by the quilt. 
4) A demonstration of autonomous quilt spreading which can handle the interference between the quilt and the human body.

\begin{figure*}[t!]
	\centering
	\includegraphics[width=\linewidth]{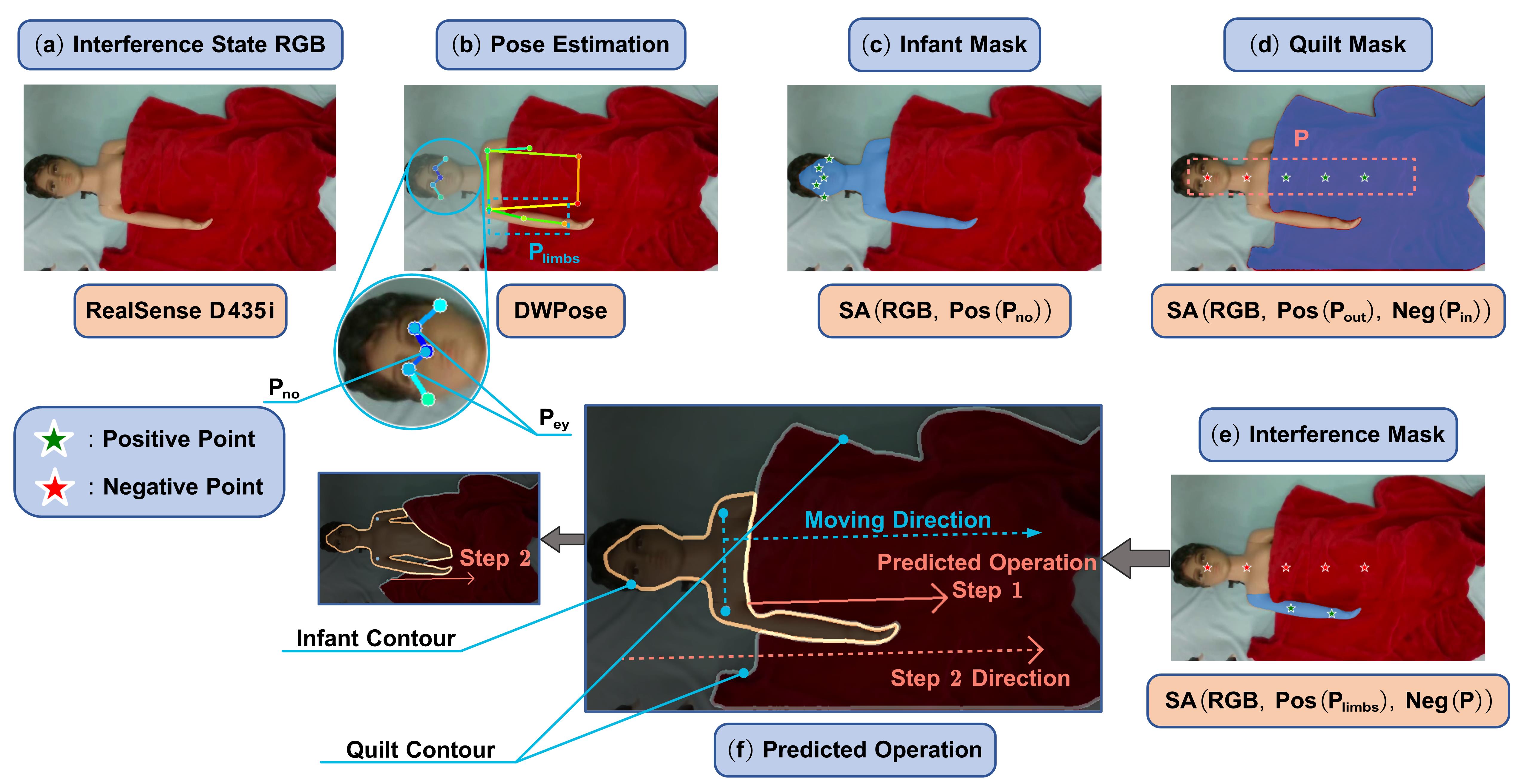}
	\vspace{-1.8em}
	\caption{Workflow for Interference Resolution. (a) shows an RGB image captured by the RealSense D435i camera. (b) presents the key human skeletal points predicted by DWPose. (c), (d), and (e) respectively illustrate the processing results for the infant mask, quilt mask, and interference mask. (f) demonstrates the operation of the interference resolution.}
	\label{interference}
	\vspace{-1em}
\end{figure*}

\section{Interference Solving}

The interference state between the infant and the quilt describes situations where the infant's body, such as arms or legs, are pressed against the quilt, hindering the functionality of inverse dynamics of the quilt which assumes the body layer and the quilt are topologically isolated. In our work, we introduce a method that recognizes and segments the interference regions using RGB images. Each detected interference is classified and resolved using a pre-defined manipulator motion based on the interference type. (see Fig. \ref{interference}(f)).

This work is focused on the most frequently occurring interference states: the infant's limbs pressing against the quilt.

\subsection{State Recognition}

In this work, recognition of the interference state is realized by combining image based segmentation (Segment Anything instance segmentation, \cite{SegmentAnything}) and human skeletal detection DWPose \cite{EffectiveWholebody}). Conventional image segmentation techniques like Mask-RCNN \cite{MaskRcnn} necessitate substantial data collection and training. Our introduction of DWPose(DW) \cite{EffectiveWholebody} and Segment Anything(SA) \cite{SegmentAnything} models sidesteps the need for datasets, thereby boosting this work's generality.

After processing the RGB image with DWPose, the key human skeletal points are extracted. By employing the nose \(P_{\text{no}}\), eyes and ears \(P_{\text{ey}}, P_{\text{ea}}\) from the human skeletal points as positive points for the SA model, we can directly segment the infant mask, as represented in:
\begin{equation} 
	\text{Infant Mask} = \text{SA}(\text{RGB}, \text{Pos}(P_{\text{no}}, P_{\text{ey}}, P_{\text{ea}})),
\end{equation}
as shown in Fig. \ref{interference}(c), where \textit{Pos} means positive points.

Building upon this infant mask, we subsequently segment the quilt mask (see Fig. \ref{interference}(d)). Point sampling for this purpose begins from the nose and extends towards the feet, following the line perpendicular to the segment connecting both shoulders. Sampling is conducted at pre-defined intervals. By referencing with the infant mask, the obtained points are classified into two sets: \( P_{\text{in}} \) (points within infant mask) and \( P_{\text{out}} \) (points outside infant mask). Together, these subsets constitute the point set \( P \). From this, the quilt mask becomes:
\begin{equation}
	\text{Quilt Mask} = \text{SA}(\text{RGB}, \text{Pos}(P_{\text{out}}), \text{Neg}(P_{\text{in}})).
\end{equation}

If limb points are detected from DWPose processing, it indicates the potential existence of interference. The exact existence can be confirmed, after the following two cases can be excluded:

1. Sometimes the detected limb points are located inside the quilt mask. This case is checked and excluded.

2. The detected limb points lie outside the quilt mask. In this case, the interference is determined by evaluating how much the limb mask area is encased by the quilt mask area as demonstrated in Fig. 3(e). If the envelopment amount is less than a threshold, it should not be recognized as interference. The evaluation is done by a module which calculates the envelopment amount of a convex hull by another one. The interference mask can be defined as:

\begin{equation}
	\text{Interference Mask} = \text{SA}(\text{RGB}, \text{Pos}(P_{\text{limbs}}), \text{Neg}(P)),
\end{equation}
where the limb points are used as positive prompts and the combination of points inside the infant and quilt mask are used as negative prompts. 

\subsection{Interference Resolution}

Utilizing the derived masks of both the quilt and the infant, as well as the infant's skeletal keypoints, we devise a strategy to address the interferences. The resolution follows this sequence:

Initially, the interference's location, either upper or lower limbs, is determined, for each case of limb interferences a pre-defined resolving strategy is applied. For upper limb interference, the approach is to drag the quilt from both sides of the arm, moving downwards to the lowest position of the visible arm until the interference is negated, as shown in Fig. \ref{interference}(f). Conversely, for lower limb interference, the strategy is to grasp the cloth nearest to the interference and move it vertically until the interference is resolved.

\section{Quilt Spreading Action Planning}
After solving the interference, the body and the quilt are topologically isolated. This means that any subsequent operation will be performed ensuring no interference between the quilt and the infant. Manipulation planning of the free quilt is done by utilizing EM*D \cite{FastFlexible}. 

\begin{figure}[t!]
	\centering
	\includegraphics[width=\linewidth]{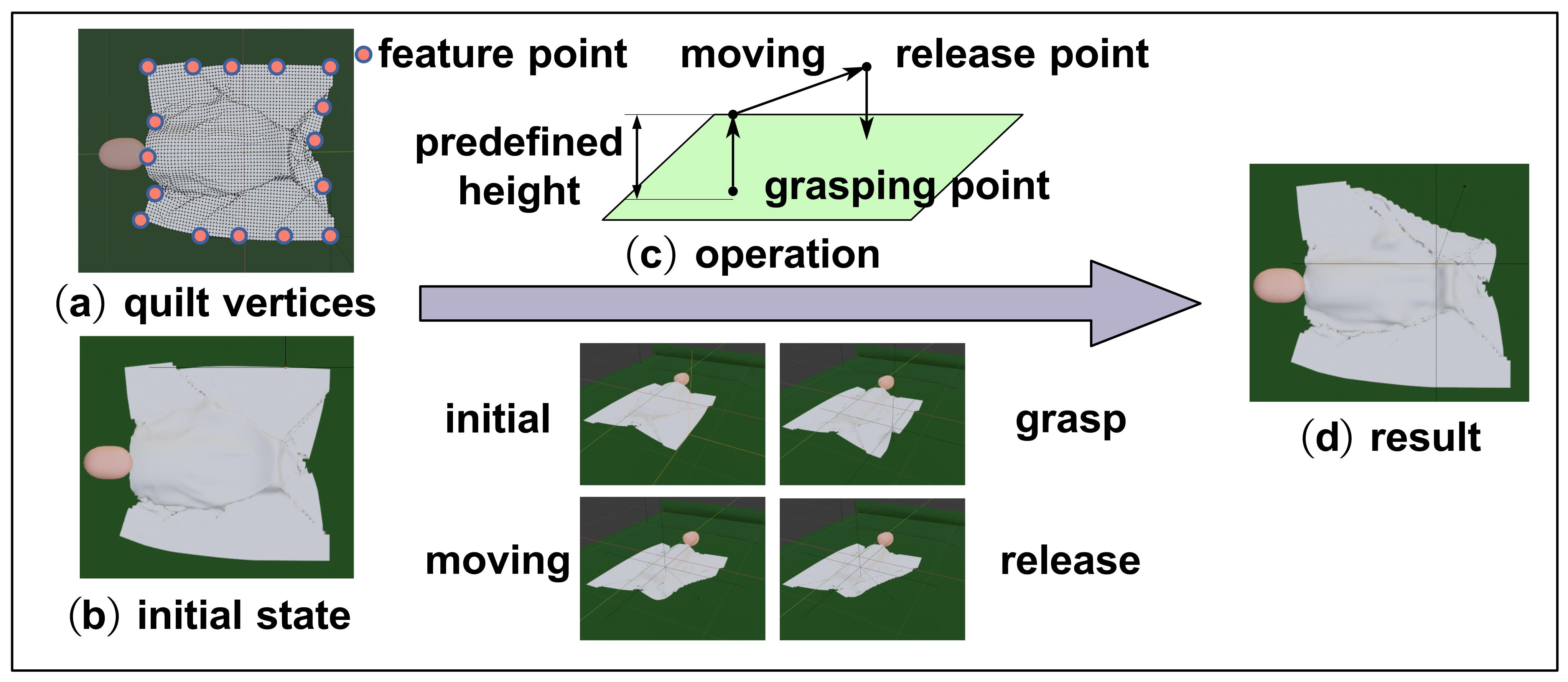}
	\vspace{-1.5em}
	\caption{Illustration of the dataset collection process: (a) the vertex representation of the quilt, (b) the initial quilt state, (c) a single quilt manipulation step, and (d) the post-manipulation quilt state.}
	\label{dataset}
	\vspace{-2em}
\end{figure}

\subsection{Dataset Acquisition}

Blender \cite{blender}, is utilized for generating the necessary data to learn the dynamics of quilt. Herein, the whole surface of a quilt is discretized into an elevation map projected to a \(64 \times 64\) grid, as visualized in Fig. \ref{dataset}(a). To mimic real-world physics, attributes like self-collision and object-collision are meticulously configured. Given the constraint on vertex distance, our experimental setup chose a comparatively thinner quilt representation.

For data acquisition, each manipulation action, denoted by \(\vec{o_{k}}\), is defined as the vector connecting the grasp point (\( g = (g_x, g_y) \)) to the release point (\( r = (r_x, r_y) \)). The grasping is executed to the vertex point corresponding to \(g\), whereas releasing is executed at a predefined altitude.

The compiled dataset is an amalgamation of such manipulative iterations, where each iteration results in a transition \((\mathbf{S_k}, \vec{o_k}, \mathbf{S_{k+1}})\) – with \(\mathbf{S_k}\) and \(\mathbf{S_{k+1}}\) symbolizing the states before and after manipulation (see Fig. \ref{dataset}(a) and (d)). Further, \(m_k = [\vec{g}, \vec{r}]\) denotes the manipulation vector (Fig. \ref{dataset}(c)). Dataset augmentation techniques including mirroring and rotation are employed, culminating in an enriched dataset of 100,000 samples. Data is collected by randomly selecting both the grasping and release points. Starting from a fully spread quilt, four sequential grasping/releasing actions are executed randomly, and the before-and-after states of the quilt, along with the action vectors, were recorded. However, the objective of our task is to spread the quilt to cover the infant with a satisfied coverage, the aforementioned automatic data collection method will never include data of the final fine adjusting phase - spreading the quilt to finally achieve good coverage state.

To address this challenge, we collected additional data, emulating the final phase of quilt-covering process. Initially, corner and edge points (see Fig. \ref{dataset}(a)) of a well spread quilt were recorded in Blender as the initial state. Then, these points were selected at random and dragged in a random direction for a variable distance. Subsequently, these points were dragged back to their initial recorded positions, serving as the generated quilt spreading data example. Finally, using visualization techniques, we manually filtered out obviously unreasonable quilt-covering actions. In total, 2,639 samples of the quilt-covering process were generated to constitute our quilt-covering dataset.


\subsection{Application of EM*D network and its Revision}
The EM*D network describes the forward dynamics of the quilt. It can be formulated as the process of deducing the resultant state \( \mathbf{S_b} \) from a given initial state \( \mathbf{S_a} \) and a sequence of manipulations \( P_{ab} = (\vec{o_1}, \vec{o_2}, \ldots, \vec{o_n}), \vec{o} \in \mathcal{O} \). This work endeavors to refine the EM*D model, ensuring compatibility with our targeted tasks. The EMD model consists of three key modules:

1. Encoder Module:
\begin{equation}
	\mathcal{E}(\mathbf{S_i}) = \mathbf{c_i}
\end{equation}

2. Manipulation Module:
\begin{equation}
	\mathcal{M}(\mathbf{c_i}, \vec{o_i}) = \hat{\mathbf{c}}_{i+1}
\end{equation}

3. Decoder Module:
\begin{equation}
	\mathcal{D}(\hat{\mathbf{c}}_{i+1}) = \mathbf{\hat{S}_{i+1}}
\end{equation}

Where:

- \( \mathbf{S_i} \): Original fabric state of dimension \( \mathbb{R}^{64 \times 64 \times 32} \).

- \( \mathbf{c_i} \): Latent representation of fabric state, dimension \( \mathbb{R}^{1 \times 1024} \).

- \( \vec{o} \): Operation vector, dimension \( \mathbb{R}^{1 \times 4} \).

- \( \hat{\mathbf{c}}_{i+1} \): Predicted latent state for the next timestep.

- \( \mathbf{\hat{S}_{i+1}} \): Predicted fabric state.


\begin{table}[t!]
	\caption{Training Parameters}
	\label{tab:parameters}
	\centering
	\vspace{-1em}
	\begin{tabular}{|l|l|}
		\hline
		Parameter & Setting \\
		\hline
		GPU Model & GTX 1080Ti × 8 \\
		\hline
		Training Set Size & 90000(random spreading) + 1631(final covering) \\
		\hline
		Test Set Size & 10000(random spreading) + 408(final covering) \\
		\hline
		Batch Size & 32 \\
		\hline
		Momentum & 0.9 \\
		\hline
		Weight Decay & 0.001 \\
		\hline
		Learning Rate & 0.00005 \\
		\hline
		Max Iterations & 3000 \\
		\hline
	\end{tabular}
	\vspace{-2em}
\end{table}

\begin{figure}[t!]
	\centering
	\includegraphics[width=\linewidth]{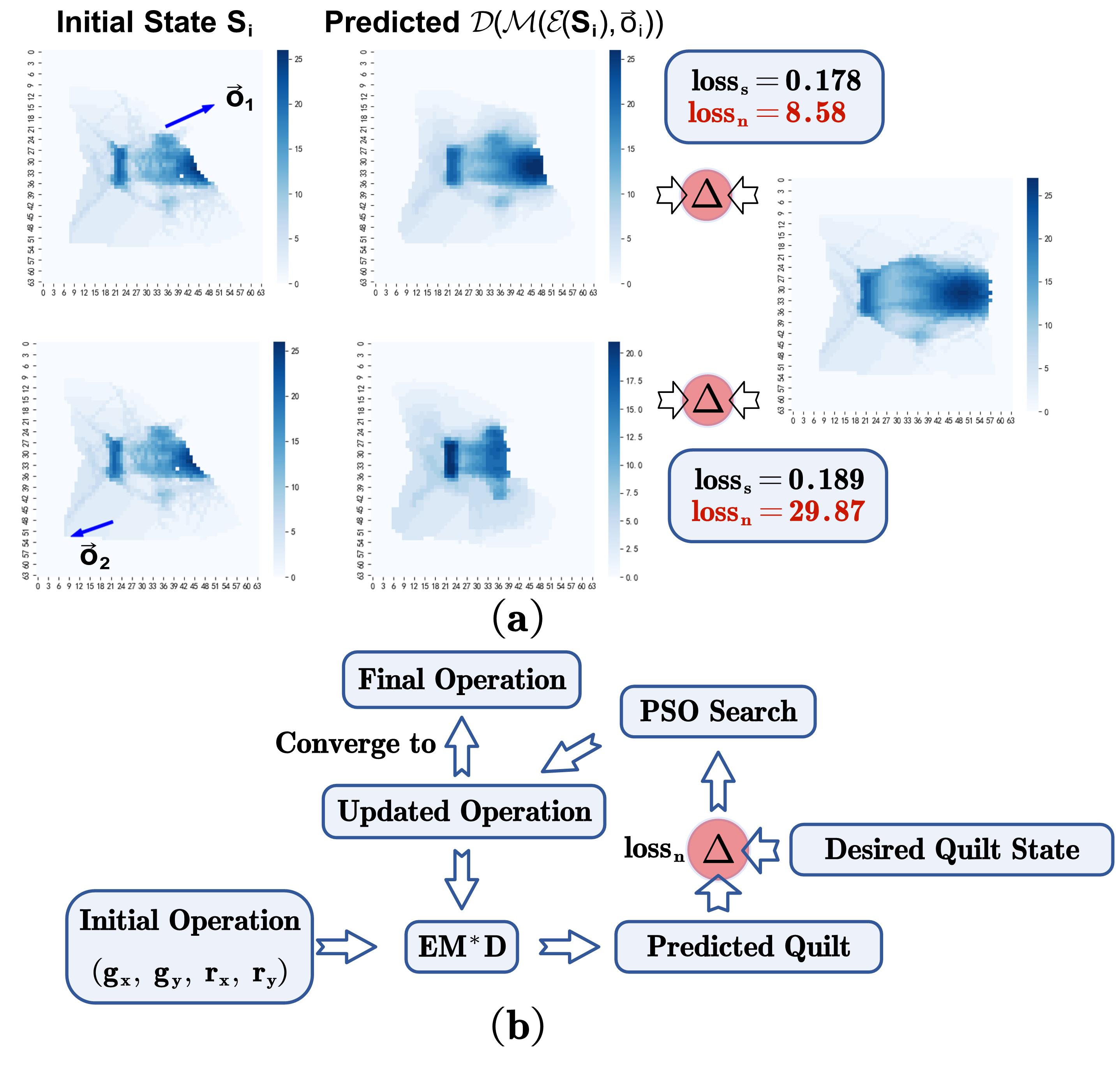}
	\vspace{-1.5em}
	\caption{Comparison of loss functions and their implications on the quilt-covering strategy. (a) Demonstrates the disparity in the performance of two loss functions, with the top row representing a rational quilt-covering approach and the bottom indicating an irrational one. (b) Solving planning problems using conventional search.}
	\label{loss}
	\vspace{-1.5em}
\end{figure}

\begin{figure}[t!]
	\vspace{1em}
	\centering
	\includegraphics[width=\linewidth]{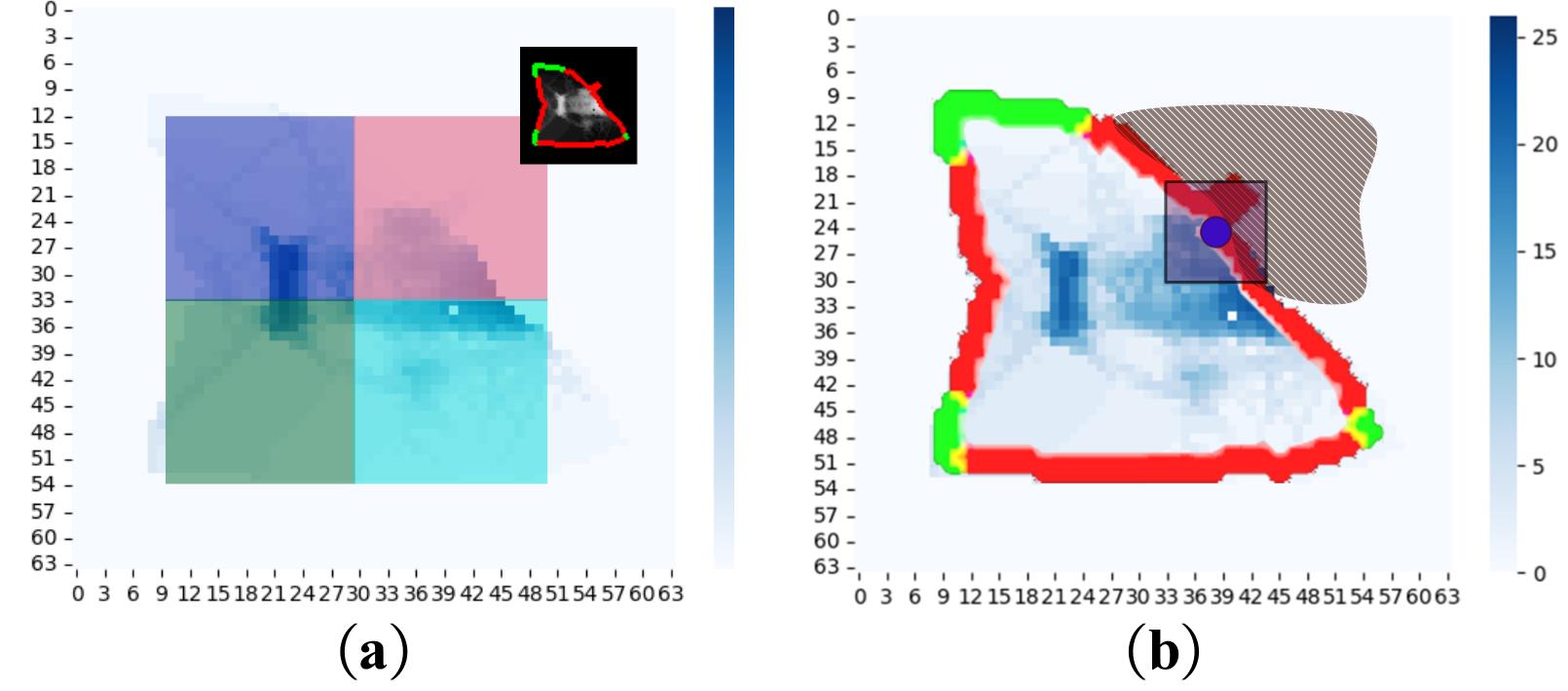}
	\vspace{-1.5em}
	\caption{Illustration of the proposed pruning strategy for grasp vector determination. (a) Segregation of the quilt domain into four distinct sections with depth image mapping to grayscale range. (b) Identification of low coverage regions in the quilt, guiding precise release point selection. }
	\label{cut}
	\vspace{-1em}
\end{figure}

Training parameters for the EM*D network are summarized in Table~\ref{tab:parameters}.

\subsection{Spreading Action Planning}
As for the spreading action planning, we adopt the method proposed in \cite{FastFlexible}. It utilizes the well trained forward dynamics model with spreading action also embedded as parameters of the model. Then action planning is formulated as optimization for the parameters concerning about spreading action. In \cite{FastFlexible}, the optimization is treated as a BP learning with the spreading action concerned components as the parameters to be tuned. Although this way will benefit from the accurately calculated gradient provided by BP back propagation mechanism, it seem apt to fall into local minimum which results in an non-optimal spreading action.  

In order to improve the accuracy of each planned spreading action and make the output of the network converge to the desired state as much as possible, two revisions are introduced. The first revision is the adoption of a new loss function.

\begin{figure}[t!]
	\centering
	\vspace{0.5em}
	\includegraphics[width=\linewidth]{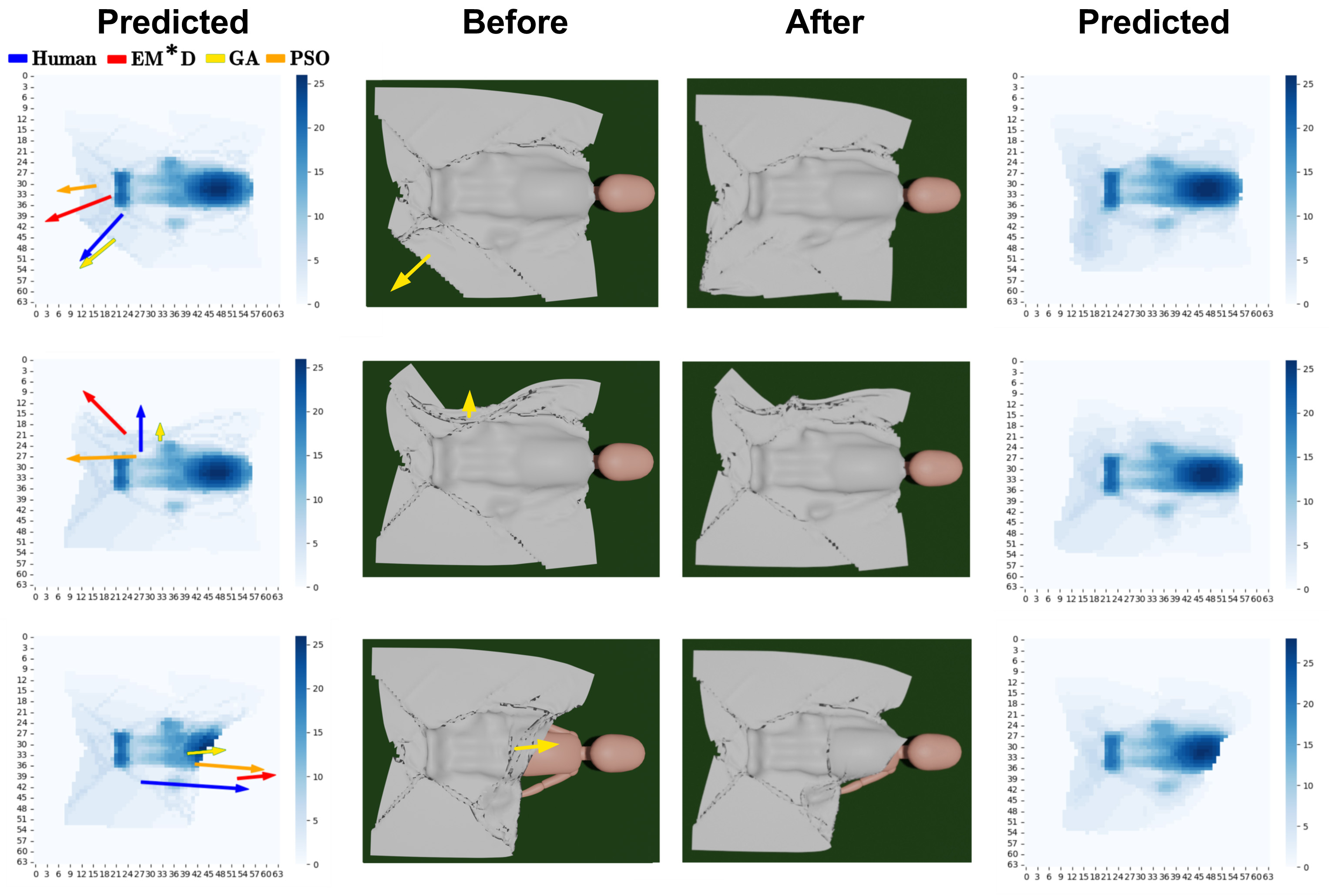}
	\vspace{-1.5em}
	\caption{Comparative results of single-step planning: optimal fabric manipulations as per human intuition (``Human'') juxtaposed against those proposed by BP style planning, Genetic Algorithm (``GA''), and Particle Swarm Optimization (``PSO'') within the Blender simulation environment.
	}
	\label{experiments1}
	\vspace{-1.2em}
\end{figure}

\begin{figure}[t!]
	\centering
	\includegraphics[width=\linewidth]{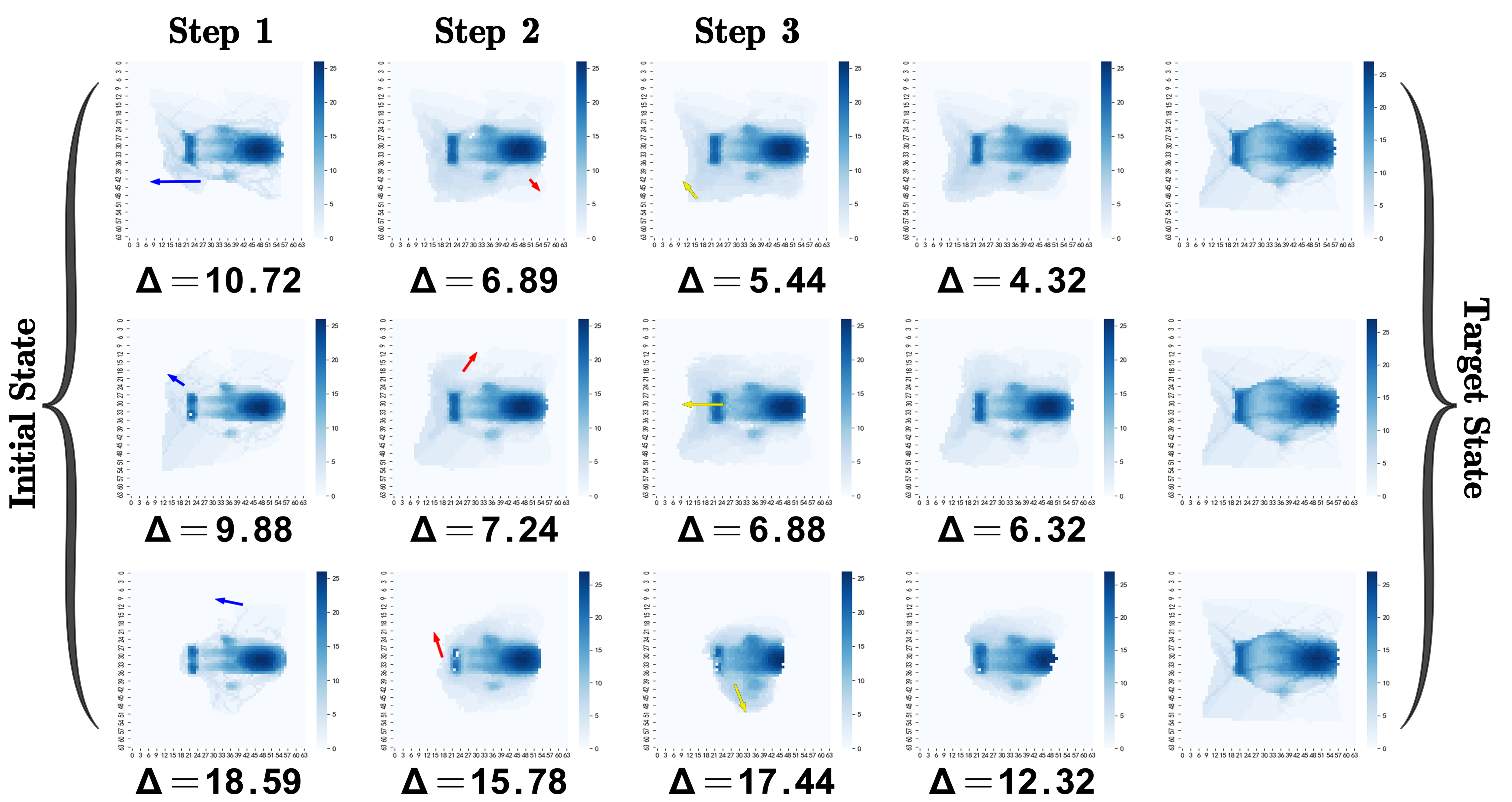}
	\vspace{-1.5em}
	\caption{Outcomes of multi-step planning: demonstration of a three-step planning approach, where the GA's performance exhibits consistent loss function reduction across datasets.
	}
	\label{experiments2}
	\vspace{-1.2em}
\end{figure}

In the initial EM*D network experiments, it was observed that the discriminative capability of the original loss function for the task was inappropriate. As illustrated in Fig. \ref{loss}(a), the first row depicts a plausible quilt spreading action, whereas the second row presents an evidently irrational one. The loss difference resulted from the two actions is negligible.

To address this problem, the original loss function is revised. In the original loss function, for each point on a fabric its altitude (the third dimension) is encoded as a length 32 binary code where ``1'' is located at the place corresponding to its altitude. In the revised function, the altitude is encoded as the depth, the loss function is computed as:  

\begin{equation}
	\text{loss}_n = \frac{1}{N} \sum_{i} \| \mathcal{D}(\mathcal{M}(\mathcal{E}(\mathbf{S_i}^{\text{depth}}), \vec{o_i})) - \mathbf{S}_{i+1}^{\text{depth}} \|^{2},
\end{equation}
where N denotes the dimension of \(\mathbf{S}^{\text{depth}}\), which is \(64 \times 64\). The second revision is the replacement of BP based searching with conventional searching technology of GA. Although search process using GA will not be directed by the calculated gradient, practically it is more superior in mitigating the local minimum.

Following human experience in quilt covering, we further designed a strategy to narrow down the search range. As shown in Fig. \ref{cut}(a), the quilt is divided into four different areas. When a particular area has a significantly lower quilt coverage than others, the weighting for release point search in that area is increased. We extract the quilt's contour and restrict grasp points to within this contour, as depicted in Fig. \ref{cut}(b). The core mathematical essence of this problem is described as follows:

\begin{equation}
	\begin{aligned}
		& \min_{\vec{o_i}} \frac{1}{N} \sum_{i} \| \mathcal{D}(\mathcal{M}(\mathcal{E}(\mathbf{S_i}^{\text{depth}}), \vec{o_i})) - \mathbf{S}_{\text{desired}}^{\text{depth}} \|^{2} \\
		& \text{s.t.} \\
		& \mathcal{F}(g_x, g_y) <= 0 \\
		& 0 \leq r_x, r_y \leq 63
	\end{aligned}
\end{equation}

- \(\mathbf{N}\): The dimension of \(\mathbf{S}^{\text{depth}}\), which is \(64 \times 64\).

- \(\mathbf{S_i^{\text{depth}}}\): A \(64 \times 64\) matrix representing the depth map at the ith step, with each element in the range \(0 \leq \mathbf{S_i^{\text{depth}}}(x, y) \leq 31\).

- \(\mathbf{S}_{\text{desired}}^{\text{depth}}\): A \(64 \times 64\) matrix representing the desired depth state of the cloth, with elements ranging from 0 to 31.

- \(\vec{o_i}\): A \(1 \times 4\) vector denoting the operation vector, encompassing the grasp point \((g_x, g_y)\) and release point \((r_x, r_y)\).

- \(\mathcal{F}(x, y)\): A function representing the interior of the quilt. When the point lie inside the quilt's contour, its value is less than 0; when on the contour, it is exactly 0; otherwise, it returns a value greater than 0.

Integrating this pruning strategy, the efficiency of the search algorithm witnessed substantial improvements. The runtime of the genetic algorithm was reduced from an initial 5 minutes to approximately 10-20 seconds, paving the way for multi-step planning.

\section{Experiments}
This section presents the performance of two manipulation action planning methods: the BP style planning and the GA Search designed under the new loss function. Both methodologies were evaluated tasks with identical initial and target states. Additionally, we validated the planned quilt spreading sequence and their effects within the simulator. In a practical setting involving interferences, we conducted experiments including interference resolution and autonomous quilt spreading following the generated plan.

\begin{figure}[t!]
	\centering
	\includegraphics[width=\linewidth]{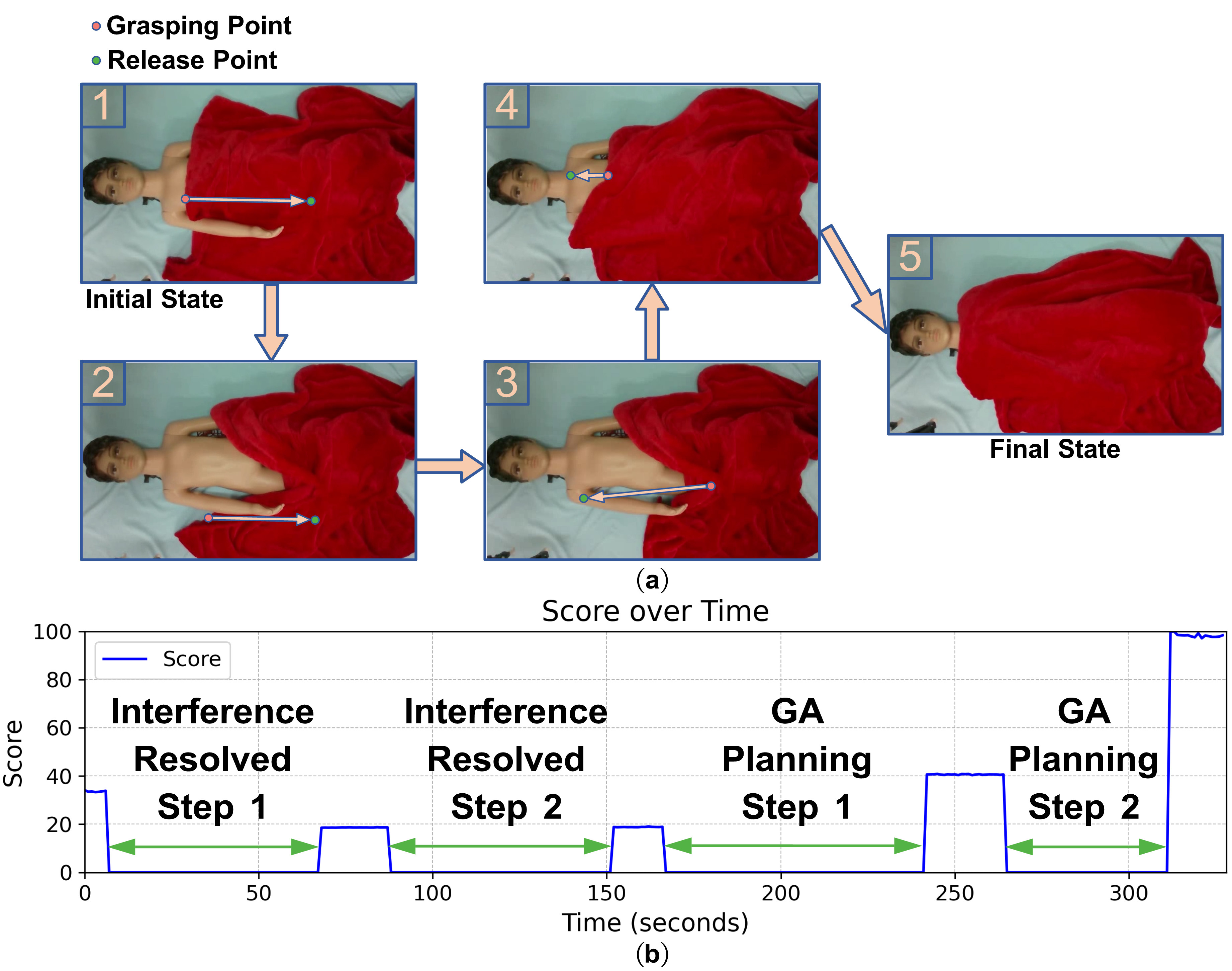}
	\vspace{-2em}
	\caption{Experimental results showcasing the quilt positioning process and its corresponding scoring system. (a) Illustrates the five stages of the real-world experiment, from identifying interference to achieving the intended quilt position. (b) Depicts the change curve of the score throughout the process, highlighting the initial score reduction due to interference resolution, followed by a systematic increase as the infant is strategically covered using the Genetic Algorithm, culminating in a scenario where only the infant's head is exposed. During the robotic arm's operation periods, the scoring system outputs 0.
	}
	\label{experiments_real}
	\vspace{-1em}
\end{figure}

\begin{figure}[t!]
	\centering
	\includegraphics[width=\linewidth]{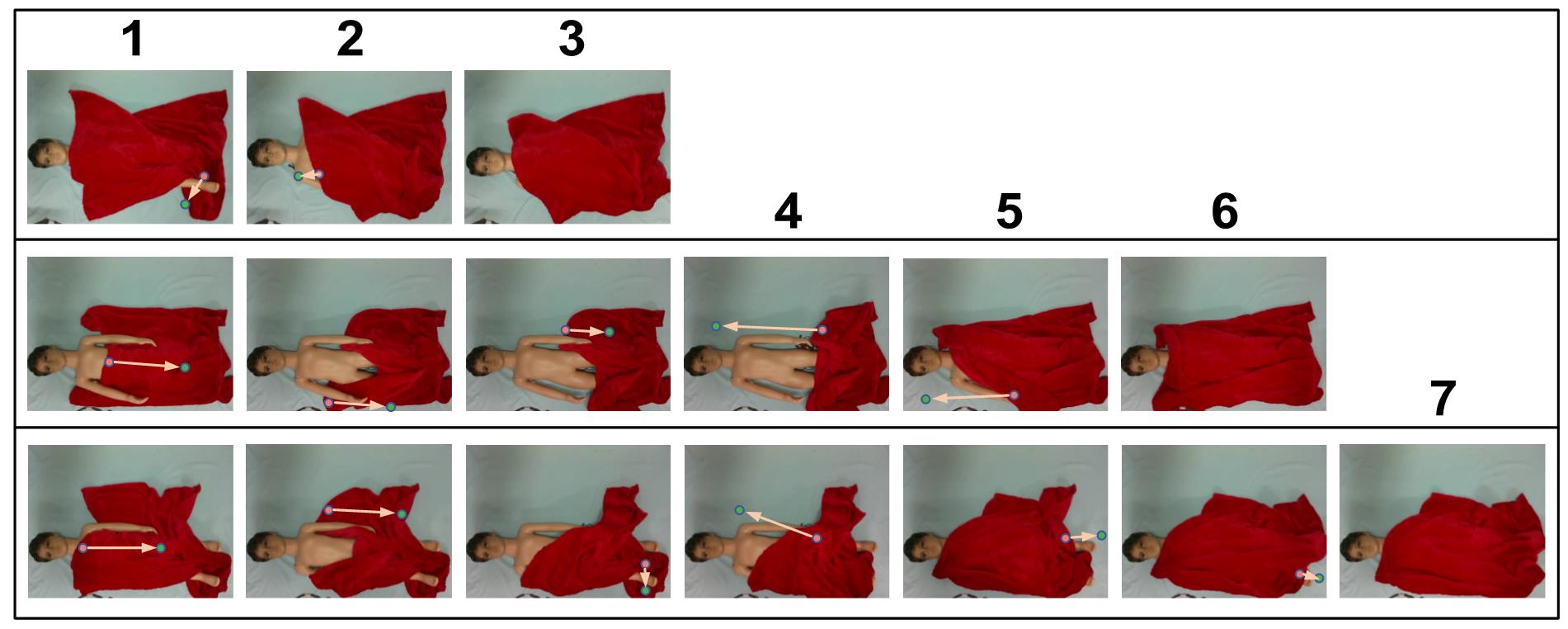}
	\vspace{-1.5em}
	\caption{Three sets of experiments demonstrated the strategy of solving interference and spreading the quilt  to cover the infant when the infant's: 1. right foot is pressing on the quilt, 2. both arms are pressing on the quilt, and 3. right foot and left arm are both pressing on the quilt.
	}
	\label{experiments_real_new}
	\vspace{-2em}
\end{figure}

\subsection{Single-step Planning in Simulation}
Fig. \ref{experiments1} displays the optimal solutions deemed by human and those provided by Searching with BP network. In the experiment using conventional search, based on practical requirements, we opted for GA and PSO. Planning vectors from these methodologies were applied to fabrics within Blender, facilitating the observation and comparison of experimental outcomes. Notably, the GA's results aligned most closely with human demonstration. In one experiment depicted in the third row, the BP search based method generated an irrational grasp-air action. The results obtained from GA search show that the state resulted from it closely match actual operations in the simulation. After test with various samples, it was deduced that the conventional reliably produces reasonable action vectors in most scenarios.

\subsection{Multi-step Planning Results}
The three-step quilt spreading generated by BP network based searching largely lacks practical applicability, often necessitating multiple attempts to yield an acceptable action. This is because due to the modeling error in forward dynamics, successive execution of manipulation actions will lead to unreasonable actions. This is often reflected in a reasonable drag in the initial step, followed by problematic actions, such as grasping at air, in the subsequent steps. In contrast, the Genetic Algorithm demonstrated commendable efficacy, as depicted in Fig. \ref{experiments2}. The results reveal that the three-step search exhibits commendable performance in the first two examples, with a consistent reduction in the loss function. However, in the third example, there exists a pronounced disparity between the final state and the target state, implying that more steps are needed for closer coincidence the target state. The search planning time for all examples hovered around one minute.

\subsection{Practical Experiment}
In a real-world experimental setting, we showcased an instance involving interference from the right arm, as shown in Fig. \ref{experiments_real}(a). Initially, our algorithm identified the existing interference and proposed an action to solve it, using the strategy illustrated in Fig. \ref{interference}. The first two steps are predominantly focused on resolving the interference. By the third step, the interference between the infant's arm and the quilt was completely decoupled. We utilize an RGB-D camera to acquire the depth information of the quilt. Based on our strategy, we will continually plan the quilt spreading approach until the quilt coverage rate meets the anticipated benchmark. Then by employing a two-step planning methodology with the Genetic Algorithm, we successfully re-cover the quilt to the intended state.

We established a scoring system where a score of 100 is awarded when only the infant's head is exposed. The more the infant's body parts are exposed besides the head, the lower the score becomes. As can be observed from Fig. \ref{experiments_real}(b): during the two steps of interference resolution, the score was reduced due to the necessity of dragging the quilt downward, resulting in greater exposure of the infant. However, in the subsequent two steps planned through the Genetic Algorithm, as the interference had already been resolved, it was possible to strategically cover the infant, leaving only the head exposed. This led to an incremental score increase until it reached the maximum—where only the infant's head remains visible. Additionally, we also conducted experiments with more interference combinations as shown in Fig \ref{experiments_real_new}. Through multiple experiments, we demonstrated the feasibility of a two-step strategy for covering the infant, involving interference resolving and quilt spreading.

\section{Conclusion}
This work introduces an action planning strategy for appropriately re-covering an infant with a quilt. Initially, the state between the infant and the quilt is discerned and segmented via RGB image. The task of re-covering the quilt is then divided into two phases: interference resolution and quilt spreading. A predefined action plan is employed for interference resolution, while the EM*D network, previously trained, is utilized for action planning. The reliability of the quilt-covering scheme is validated through both simulations and practical experiments. In this work, the proposed method solely relies on RGB images for state recognition. This could lead to misjudgments, especially when the patterns or colors of the infant's clothes closely resemble those of the quilt. For practical implementation, we are considering the integration of a thermal imager to directly segment the infant's portion, thereby reducing the likelihood of misidentification.

\bibliographystyle{IEEEtran}
\balance
\bibliography{references}
\end{document}